%% file: acl_latex.tex
\pdfoutput=1
\documentclass[11pt]{article}

\usepackage[preprint]{acl}

\usepackage{times}
\usepackage{latexsym}
\usepackage{booktabs}
\usepackage{adjustbox}
\usepackage{multirow}
\usepackage{amsmath}
\usepackage[most]{tcolorbox}
\usepackage{makecell}
\usepackage{xcolor}
\usepackage{color}
\usepackage{colortbl}
\usepackage{amssymb}  %
\usepackage{amsthm}
\usepackage{acronym}
\usepackage[T1]{fontenc}

\usepackage[utf8]{inputenc}

\usepackage{microtype}

\usepackage{inconsolata}

\usepackage{graphicx}

\usepackage{subcaption} %

\usepackage{amsmath}        %
\usepackage[linesnumbered,ruled,vlined]{algorithm2e}
\usepackage[inline]{enumitem}

\newtcbox{\hlprimarytab}{on line, rounded corners, box align=base, colback=backblue, colframe=white, size=fbox, arc=3pt, before upper=\strut, top=-2pt, bottom=-4pt, left=-2pt, right=-2pt, boxrule=0pt}
\newtcbox{\hlsecondarytab}{on line, box align=base, colback=backred, colframe=white, size=fbox, arc=3pt, before upper=\strut, top=-2pt, bottom=-4pt, left=-2pt, right=-2pt, boxrule=0pt}

\definecolor{backred}{RGB}{255, 190, 190}
\definecolor{backblue}{RGB}{220, 230, 250}

\definecolor{Gainsboro}{rgb}{0.86, 0.86, 0.86}
\definecolor{Gray}{gray}{0.95}

\newtheorem{theorem}{Theorem}[section]
\newtheorem{lemma}[theorem]{Lemma}

\newcommand{\fullmodel}{\underline{\textbf{M}}ulti-agent J\underline{\textbf{o}}int \underline{\textbf{A}}lignment \underline{\textbf{T}}uning\xspace}
\newcommand{\model}{MOAT\xspace}
\newcommand{\first}{Planning Agent Alignment\xspace}
\newcommand{\second}{Grounding Agent Improving\xspace}

\acrodef{mact}[MACT]{multi-agent collaboration tuning}
\acrodef{llm}[LLM]{large language model}
\acrodef{dpo}[DPO]{direct preference optimization}

\title{\textit{Bridging the Capability Gap}: Harmonizing Multi-Agent Systems \\via Joint Alignment Tuning}

\author{
 \textbf{Minghang Zhu\textsuperscript{1}}\thanks{Equal contributions}\space\space\space
 \textbf{Zhengliang Shi\textsuperscript{1}}\footnotemark[1]\space\space\space
 \textbf{Zhiwei Xu\textsuperscript{1}}\space\space\space
 \textbf{Shiguang Wu\textsuperscript{1}}\space\space\space
\\
 \textbf{Lingjie Wang\textsuperscript{1}}\space\space\space
 \textbf{Pengjie Ren\textsuperscript{1}}\space\space\space
 \textbf{Zhaochun Ren\textsuperscript{2}}\footnotemark[2]\space\space\space
 \textbf{Zhumin Chen\textsuperscript{1}}\thanks{Corresponding authors}
\\
 \textsuperscript{1}Shandong University, Qingdao, China \\
 \textsuperscript{2}Leiden University, Leiden, The Netherlands
\\
 \texttt{ \{mhzhu,shizhl\}@mail.sdu.edu.cn} \\ \texttt{z.ren@liacs.leidenuniv.nl, chenzhumin@sdu.edu.cn}
}

\begin{document}
\maketitle
\input{sections/00-abstract}
\input{sections/01-introduction}

\input{sections/02-related-work}
\input{sections/03-method}
\input{sections/03-theory}

\input{sections/04-experiment-set-up}

\input{sections/05-experiment}

\input{sections/06-conclusion}

\section*{Limitations}

Our framework is currently developed and evaluated exclusively on text-based scenarios, without exploring multimodal learning settings. 
While modern open-source language models (e.g., LLaVA, Qwen-VL) have demonstrated emerging capabilities in processing multimodal inputs, our current architecture lacks explicit mechanisms for cross-modal alignment during collaborative training.
In future work, we plan to incorporate multimodal information into our framework.

\section*{Ethics Statement}

This research strictly adheres to the ethical principles outlined in the ACM Code of Ethics, with rigorous implementation of transparency and accountability measures. All datasets, tools, and language models (including Llama-2, Mistral and Qwen) are sourced from publicly available platforms under compliant licenses, ensuring ethical alignment and reproducibility. The complete code and evaluation protocols are open-sourced. 

\section*{Acknowledgements}

This work was supported by the National Natural Science Foundation of China under Grant No. 62372275 and 62472261, the Technology Innovation Guidance Program of Shandong Province under Grant No. YDZX2024088, the Provincial Key R\&D Program of Shandong Province under Grant No. 2024CXGC010108.

\bibliography{custom}

\clearpage
\newpage
\appendix
\input{sections/appendix}

\end{document}

%% file: sections/00-abstract.tex
\begin{abstract}

    The advancement of large language models (LLMs) has enabled the construction of multi-agent systems to solve complex tasks by dividing responsibilities among specialized agents, such as a planning agent for subgoal generation and a grounding agent for executing tool-use actions. Most existing methods typically fine-tune these agents independently, leading to capability gaps among them with poor coordination. To address this, we propose \model, a Multi-Agent Joint Alignment Tuning framework that improves agents collaboration through iterative alignment. \model alternates between two key stages: (1) Planning Agent Alignment, which optimizes the planning agent to generate subgoal sequences that better guide the grounding agent; and (2) Grounding Agent Improving, which fine-tunes the grounding agent using diverse subgoal-action pairs generated by the agent itself to enhance its generalization capability. Theoretical analysis proves that \model ensures a non-decreasing and progressively convergent training process. Experiments across six benchmarks demonstrate that \model outperforms state-of-the-art baselines, achieving average improvements of 3.1\% on held-in tasks and 4.4\% on held-out tasks.
    \footnote{Code is available on \url{https://github.com/ZMingHang/MOAT/tree/master}.} 
\end{abstract}

%% file: sections/01-introduction.tex
\section{Introduction}\label{sec:intro}

\begin{figure}[!t]
        \centering
	\includegraphics[width=1 \linewidth]{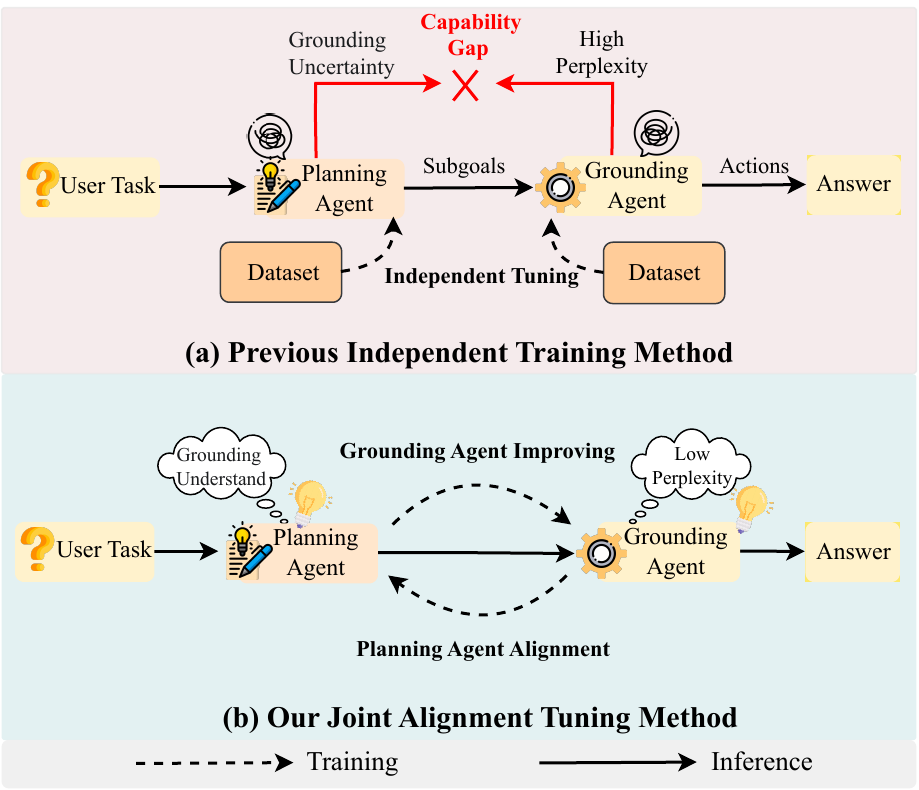}
        \caption{Comparison between (a) previous independent training method and (b) our joint alignment tuning method MOAT. The MOAT performs iterative joint tuning to align agent capabilities and improve coordination.}
 \label{fig:intro}
\end{figure}

The rapid advancement of large language models (LLMs) has significantly transformed the development of intelligent agents capable of reasoning, decision-making, and interacting with complex environments~\cite{sumerscognitiveagengt,llmplanner,ChaseLangChain2022}. Previous work typically involves prompting or fine-tuning a single foundation model on a specific dataset, training the LLMs how to use external search engines for information retrieval or call Web APIs for tasks like travel planning~\cite{qin2024toolllm,xie2024travelplanner,song2023restgpt}. 
Recently, to enable LLM-based agents to handle more real-world and multi-step tasks, more and more research has increasingly focused on multi-agent systems~\cite{lumos,wang2025cooperative,shen-etal-2024-small,wang2024peerexpert,chen2025brainstormingdriveshighqualityscientific}, which aim to synergize functionality-specialized agents. 
Figure~\ref{fig:intro} illustrates a commonly-used pipeline, where a multi-agent system typically includes a planning agent that decomposes the task into subgoals, followed by a grounding agent that executes these subgoals by invoking appropriate tools, ultimately producing the final solution.

Despite the progress made by existing multi-agent systems, effectively aligning different agents toward holistic performance remains an active research challenge. Most existing methods construct specific training data for each agent and train each agent independently. While this decomposition can enhance overall performance, it does not guarantee effective collaboration among agents.
As illustrated in Figure~\ref{fig:intro}(a), independently trained agents often exhibit varying levels of proficiency, leading to capability mismatches. For example, a planning agent might generate high-level subgoals that are difficult for a weaker grounding agent to understand and execute. Conversely, a strong grounding agent might struggle with subgoals generated by a weaker planning agent, leading to errors or inefficient task execution. Without explicit mechanisms for adapting to each other's behaviors, these agents struggle to collaborate effectively, resulting in misaligned interactions and coordination failures.
In this work, we focus on this plan-ground-execute paradigm since it has been widely adopted in most multi-agent frameworks~\cite{lumos,shen2024small,shi-etal-2024multi-agent_tool_learning}, serving as a general and foundational pipeline.

To address the above challenges, we propose \model, a \fullmodel framework that iteratively alternates between two key stages to achieve alignment in a multi-agent system: (i) \textbf{\first}, and (ii) \textbf{\second}.
Unlike prior works that train each agent independently, MOAT performs multi-agent joint alignment tuning by iteratively and coordinately optimizing the planning and grounding agents. 

Specifically, in the \first stage, \model optimizes the planning agent to generate subgoals that better guide the grounding agent in producing correct tool-calling actions. 
Given an input task, we first sample multiple candidate sequences of subgoals from the planning agent.
For each subgoal sequence, we then evaluate its effectiveness by measuring the perplexity of the grounding agent in generating correct tool calls conditioned on each sequence.
Perplexity reflects how well the grounding agent can follow a subgoal, where lower perplexity indicates higher suitability~\cite{gao2024confucius}.
Using this as a reward signal, we apply the direct preference optimization (DPO) algorithm~\cite{rafailov2024dpo} to train the planning agent to align with the grounding agent's preferences.
In the \second stage, we aim to enhance the grounding agent’s ability to interpret and act upon subgoals produced by the planning agent.
Specifically, we reuse the subgoal sequences from the planning agent in the first stage as training data, exposing the grounding agent to realistic settings.
For each input task, we use the subgoal–action pairs to train the grounding agent via standard language modeling loss. 
Comparing with relying on ground truth or handcrafted input subgoals, this allows the grounding agent to adapt to the distribution of subgoals produced by the planning agent at practical inference time, thereby improving its robustness and execution accuracy.

Through theoretical analysis, we demonstrate that the holistic performance of the multi-agent system is improved progressively by alternating the above two stages.
We apply \model to several open-source model families (Llama, Mistral, and Qwen) and evaluate it on three types of tasks, i.e., : Web, Math, and QA, across six benchmarks. The results show that \model consistently outperforms existing baselines,  on both in-distribution training sets and out-of-distribution test sets. These validate the effectiveness of our joint alignment framework and demonstrate its strong generalization ability.

Our main contributions are as follows:
(i) We introduce \model, a multi-agent joint alignment tuning framework to jointly optimize interconnected agents, bridging the capability gap between them;
(ii) We provide formal analysis proving that the alternating optimization of planning and grounding agents guarantees non-decreasing performance and convergence; and 
(iii) Extensive experiments on both held-in and held-out settings across six benchmarks demonstrate that \model achieves the best performance with 4.4\% improvement.

%% file: sections/02-related-work.tex
\section{Related work}

\paragraph{LLM-based multi-agent system.}

Large language models (LLMs) have enabled the development of autonomous agents capable of reasoning, planning, tool use, and memory retention to solve specific goals through self-directed interaction and decision-making~\cite{liuagentbench,madaan2024selfreflection,lyu2024knowtuning}.
These agents have demonstrated strong capabilities across various complex tasks, such as web navigation \cite{yao2022webshop, zhou2023webarena}, task planning \cite{zhang-etal-2024-exploring}, and tool learning \cite{shi-etal-2024multi-agent_tool_learning}.
While single-agent frameworks like AutoGPT~\cite{yang2023autogpt}, XAgent~\cite{xagent2023}, and LangChain~\cite{ChaseLangChain2022} address such tasks by equipping a single LLM agent with external tools and functions, recent work has explored multi-agent systems that improve problem-solving efficiency through collaborative interaction among multiple agents.
For example, CAMEL~\cite{li2023camel}, AutoGen~\cite{wu2024autogen}, MetaGPT~\cite{hongmetagpt}, and ChatEval~\cite{chan2024chateval} employ role-playing and structured dialogues to improve task-solving efficiency. However, these systems typically rely on closed-source models, limiting their transparency and practical deployment in privacy scenarios.

\paragraph{Agent tuning.} 

Agent tuning improves a model’s ability to perform downstream tasks by fine-tuning open-source LLMs using trajectories distilled from stronger models~\cite{song2024agentbank,chen2023fireact,lyu2024macpo,chen2025mpaw}. 
For example, approaches such as AgentTuning~\cite{zeng-etal-2024-agenttuning}, and AgentOhana~\cite{zhang2024agentohana} fine-tune smaller models on datasets generated by GPT-series LLMs. While these improve instruction following and reasoning, single-agent tuning remains limited for complex tasks requiring long-term planning and execution~\cite{liuagentbench}.
To overcome these limitations, frameworks like Lumos~\cite{lumos} and $\alpha$-UMi~\cite{shen-etal-2024-small} propose multi-agent training methods that enable collaboration across functionality-specialized agents.
More recent work like AutoACT~\cite{qiao2024autoact} further advances this direction by introducing a self-training process where each agent is trained on a dataset generated by itself.
However, existing methods often train agents independently, lacking joint optimization to ensure effective coordination.
In contrast, our work performs iterative joint tuning to align agents' capabilities for improved cooperation.

%% file: sections/03-method.tex
\section{Task Preliminary}\label{sec:preliminary}

A multi-agent system typically consists of three components: (1) a planning agent that breaks down tasks into subgoals, (2) a grounding agent that converts subgoals into executable actions, and (3) an execution module that carries out the actions to get the final answer.
{Given a complex task $x$, the planning agent, denoted as $\pi_{p}$ is tasked to decompose it to a sequence of subgoals, formulated as $\boldsymbol{s} =\pi_{\mathrm{p}}(x) = \{s_i \mid i \in [|\boldsymbol{s}|]\}$.
Each \( s_i \) represents a subgoal like \texttt{''Calculate the total number of units in the entire building''}, contributing to solving  the overall task \( x \).
Next, the grounding agent, denoted as \( \pi_{g} \), takes the task \( x \), the set of available tools \( I \) as well as the decomposed subgoals \( \boldsymbol{s} \) as input to generates a sequence of tool calls \( \boldsymbol{a} = \pi_{\mathrm{g}}(x, I, \boldsymbol{s}) = \{a_i \mid i \in [|\boldsymbol{a}|]\} \).
Each \( a_i \in I \) represents an individual tool invocation required to complete the subgoal \( \boldsymbol{s}_i \), such as ``\texttt{R1 = Calculator(15 * 8)}''.
Finally, the execution module is responsible for executing the generated tool-call sequence \( \boldsymbol{a} \) to accomplish the user task \( x \).}

\begin{figure*}[htbp]
        \centering
	\includegraphics[width=1\linewidth]{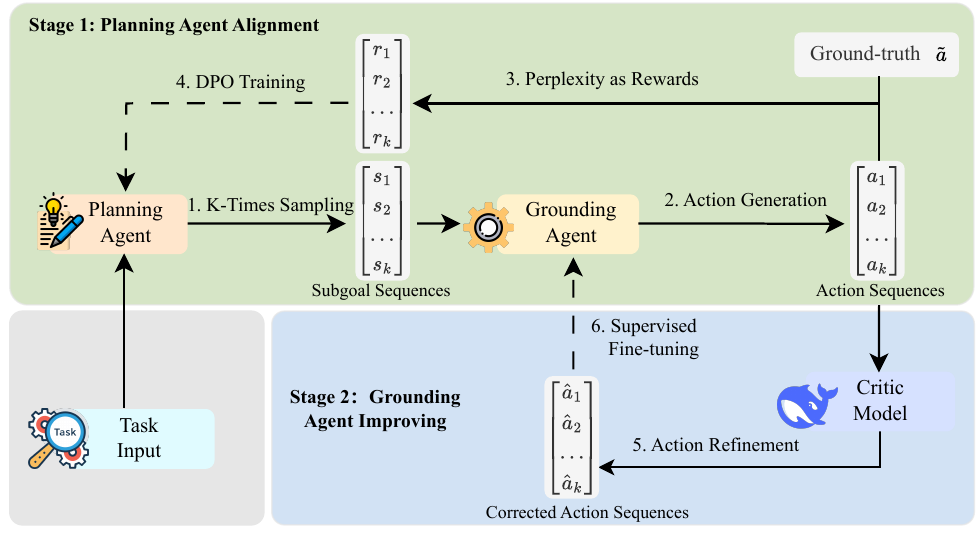 }
        \caption{
        The proposed \model framework iteratively alternates between two stages: (1) \first: The planning agent samples $K$ candidate subgoal sequences, and the grounding agent generates corresponding tool-calling actions. Subgoal sequences are ranked by PPL, and the planning agent is optimized via DPO; (2) \second: The subgoal-action pairs generated are corrected using a critic model, and the grounding agent is fine-tuned on the corrected dataset to enhance generalization.
        }
        \label{fig:method}
\end{figure*}

\section{Multi-agent Joint Alignment Tuning}\label{sec:method}

The proposed \textbf{multi-agent joint alignment tuning} (\model) framework aims to iteratively align the planning and grounding agents, enhancing the overall performance and coordination of the multi-agent system.
As illustrated in Figure~\ref{fig:method}, \model alternates between two stages: 
(1) \textit{\first}, where the planning agent explores diverse subgoal sequences to guide the grounding agent more effectively, and
(2) \textit{\second}, where we reuse the generated sub-goal sequences, improving the grounding agent to better understand them and generate correct actions.

Through this iterative process, both agents progressively adapt to each other, resulting in more coherent subgoal generation, and more accurate tool calling, towards holistic improvement.

\subsection{\first}

As illustrated by previous work~\cite{sun2023contrastive,shi-etal-2024-generate,hou2025advancingreason}, LLMs have encoded strong knowledge and reasoning abilities in their parameter space, which enables them to generate diverse and meaningful subgoal sequences through sampling.
However, not all sampled subgoal sequences are equally effective—some better guide the grounding agent to generate correct tool-use actions.
To exploit this potential, we sample multiple subgoal candidates and optimize the planning agent to prefer those that lead to better grounding outcomes.

Given a task \(x\), we sample \(K\) candidate subgoal sequences from the planning agent as \( \boldsymbol{s} = \pi_{\mathrm{p}}(x)\) and obtain a set \(\mathcal{S} = \{\boldsymbol{s}_{j} \mid j \in [|K|]\}\).
For each \(\boldsymbol{s} \in \mathcal{S}\), we calculate its perplexity (PPL) with respect to the grounding agent, where a lower perplexity indicates that the subgoal sequence is more helpful to the grounding agent, facilitating the generation of correct responses. 
Therefore, the PPL can directly reflect how \(\boldsymbol{s}\) is useful to the end-to-end task performance, which is formulated as follows:
\begin{equation}
\small
    \begin{aligned}
    \label{eq:ppl}
    &\mathrm{PPL}_{\pi_\mathrm{g}}(\boldsymbol{a} \mid x,I,\boldsymbol{s}) := \\
    & \exp\Big\{- \frac{1}{|\boldsymbol{a}|}\sum_{i=1}^{|\boldsymbol{a}|}\log P_{\pi_\mathrm{g}}(\boldsymbol{s} \mid \boldsymbol{a}_{< i}, x, I, \boldsymbol{s})\Big\}.\nonumber
\end{aligned}
\end{equation}

To align the planning agent with holistic task-solving performance, we train the planning agent to generate subgoal sequences with lower PPL, as desirable behaviors, while penalizing undesirable ones, i.e., subgoal sequences with high PPL. 
Specifically, we adopt the Direct Preference Optimization (DPO) algorithm~\cite{rafailov2024dpo}, {which allows us to align the model by learning from preference pairs. Specifically, we construct preference pairs \((\boldsymbol{s}_w, \boldsymbol{s}_l)\) from the sampled candidates in $\mathcal{S}$, where \(\boldsymbol{s}_w\) yields the lowest perplexity (strongest grounding guidance) and \(\boldsymbol{s}_l\) the highest.}

Formally, the loss function is calculated as:
\begin{equation*}
\small
\begin{aligned}
    &\mathcal{L}_{\mathrm{DPO}} = -\mathbb{E}_{(t, (\boldsymbol{s}_w, \boldsymbol{s}_l))\sim D} \\
    &\left[ \log\sigma \left( \beta \log \frac{\pi_{\mathrm{p}}(\boldsymbol{s}_w|x)}{\pi_{\mathrm{ref}}(\boldsymbol{s}_w|x)} - \beta \log \frac{\pi_{\mathrm{p}}(\boldsymbol{s}_l|x)}{\pi_{\mathrm{ref}}(\boldsymbol{s}_l|x)} \right) \right].
\end{aligned}
\end{equation*}
Here \(\pi_{\mathrm{ref}}\) represents the reference model, which is initialized as the original \(\pi_{\mathrm{p}}\) before optimization. And \(\sigma\) denotes the sigmoid function, \(\beta\) is a hyperparameter.

\subsection{\second}

This stage aims to enhance the generalization capability of the grounding agent and improve its adaptability to the diverse subgoal sequences generated by the planning agent.
To achieve this, we reuse the diverse subgoal sequences sampled from the first stage as inputs, and prompt the grounding agent to generate outputs to fine-tune itself.
Specifically, given a task $x$, for each subgoal sequence $\boldsymbol{s}$ from the sampled set $\mathcal{S} = \{\boldsymbol{s}_j \mid j \in [|K|]\}$, the grounding agent $\pi_{\mathrm{g}}$ generates the corresponding tool-calling actions as $\boldsymbol{a}_{j} = \pi_{\mathrm{g}}(x,I,\boldsymbol{s}_{j})$.

{However, since these action sequences are model-generated, they may contain errors. Directly using such noisy data for fine-tuning can lead to performance degradation or even training collapse, as highlighted in prior work~\cite{dohmatob2024collapse,shumailov2024collapse}.
To address this issue, we introduce a validation mechanism that filters out incorrect outputs before using them for training.
In particular, we employ a more powerful LLM as a critic model to evaluate whether each generated sequence $\boldsymbol{a}$ successfully solves the task, given the input $(x,I,\boldsymbol{s})$. If the critic determines that the sequence fails to complete the task, it provides a corrected version $\hat{\boldsymbol{a}}$ by referencing the ground-truth outcome $\Tilde{\boldsymbol{a}}$.
This filtering and correction process ensures that only reliable supervision signals are used during grounding agent training. The full procedure is summarized in Algorithm~\ref{alg:dataset_construction}, and the critic prompting strategy is detailed in Appendix~\ref{app:prompt}.}

\begin{algorithm}[!t]
\SetAlgoLined
Initialize SFT dataset $\mathcal{D}_{\mathrm{g}} \leftarrow \emptyset$\;
\For{each task $x$ and $s \in \mathcal{S}$}{
    Generate $\boldsymbol{a} \leftarrow \pi_{\mathrm{g}}(x, I, \boldsymbol{s})$\;
    \If{$\textrm{Critic}(\langle x, I, \boldsymbol{s} \rangle, \boldsymbol{a}, \Tilde{\boldsymbol{a}}) = \mathrm{False}$}{
    $\hat{\boldsymbol{a}} = \textrm{Critic}(\langle x, I, \boldsymbol{s} \rangle, \boldsymbol{a}, \Tilde{\boldsymbol{a}})$\;
    
    $\mathcal{D}_{\mathrm{g}} \leftarrow \mathcal{D}_{\mathrm{g}} \cup \left\{ (x, I, \boldsymbol{s}), \hat{\boldsymbol{a}} \right\}$\;
        
    }
    \Else {
        $\mathcal{D}_{\mathrm{g}} \leftarrow \mathcal{D}_{\mathrm{g}} \cup \left\{ (x, I, \boldsymbol{s}), \boldsymbol{a} \right\}$\;
    }
}
\Return SFT dataset $\mathcal{D}_{\mathrm{g}}$\;
\caption{Dataset Construction}
\label{alg:dataset_construction}
\end{algorithm}

{The overall \model alternates between the first and second stages described above.
During this process, the planning agent gradually adapts to the grounding agent by generating subgoal sequences that better align with its inference process; the grounding agent, in turns, improves its generalization capability to understand the subgoals of the planning agent.
This formulates a loop for a consistent improvement.
We also provide a detailed pseudo algorithm in Algorithm~\ref{alg:iterative_optimization} to further clarify our joint training process.}

\subsection{Cold Start}
Before applying our joint alignment strategy, we perform cold start training to equip both agents with basic task-solving capabilities, following previous work~\cite{yuan2025selfrewardinglanguagemodels,zhu2024atm,su2025openthinkimg}.
We first conduct an initial tuning using the supervised fine-tuning (SFT) dataset collected in previous work \cite{lumos}.
Specifically, the planning agent is trained to generate the correct subgoals $\boldsymbol{s}$ for an input task $x$, formulated as:
\begin{equation}
\mathcal{L}_{\mathrm{p}} = -\sum\nolimits_{i=1}^{|\boldsymbol{s}|} \log P_{\pi_{\mathrm{p}}}({s}_{i} \mid {s}_{<i}, x),
\end{equation}
The grounding agent is optimized to ground the subgoals $\boldsymbol{s}$ to the corresponding tool-calling actions $\boldsymbol{a}$, formulated as:
\begin{equation}
\mathcal{L}_{\mathrm{g}} = -\sum\nolimits_{i=1}^{|\boldsymbol{a}|} \log P_{\pi_{\mathrm{g}}}({a}_{i} \mid {a}_{<i}; x,I,\boldsymbol{s}),
\end{equation}
where \( I \) is the list of external tools. The final answer is obtained by executing the tool-callings $\boldsymbol{a}$.

%% file: sections/03-theory.tex
\section{Theoretical analysis}\label{sec:therotical}

In our framework, the planning agent and grounding agent are optimized iteratively.
In this section, we provide a theoretical analysis to demonstrate that each optimization step leads to non-decreasing improvements and ultimately ensures the convergence.
We start by defining the expected performance of the overall multi-agent system as:
\begin{equation}
    \mathbb{E}[R] = \mathbb{E}_{\boldsymbol{s} \sim \pi_{\mathrm{p}}(x)} \left[ \mathbb{E}_{\boldsymbol{a} \sim \pi_{\mathrm{g}}(s)} [R(\boldsymbol{s}, \boldsymbol{a})] \right].
\end{equation}
Here the reward function $R(\boldsymbol{s}, \boldsymbol{a})$ evaluate the quality of tool-calling action $\boldsymbol{a}$ given sub-goal sequence $\boldsymbol{s}$.
And $x$ indicates the input task. 
Below, we can state the following two lemmas.

\begin{lemma}\label{lem:plan}
    Optimizing the planning agent while keeping the grounding agent fixed leads to a non-decreasing expected reward.
\end{lemma}
    The planning agent is optimized using DPO, with PPL as the reward signal. The optimization objective can be formalized as:
    \begin{equation}
        \max\nolimits_{\pi_\mathrm{p}} \, \mathbb{E}_{\boldsymbol{a} \sim \pi_{\mathrm{p}}(x)} \left[ -\mathrm{PPL}(\boldsymbol{a}; \pi_{\mathrm{g}}) \right].
    \end{equation}
    Since PPL is negatively correlated with the true reward $R(\boldsymbol{s}, \boldsymbol{a})$, this is equivalent to maximizing the expected reward:
    \begin{equation}
        \max\nolimits_{\pi_\mathrm{p}} \, \mathbb{E}_{\boldsymbol{s} \sim \pi_{\mathrm{p}}(x)} \left[ R(\boldsymbol{s}, \boldsymbol{a}) \right].
    \end{equation}
    The DPO algorithm guarantees that updates to $\pi_{\mathrm{p}}$ lead to non-decreasing expected rewards when the grounding agent is fixed. Thus, we have:
    \begin{equation}
        \mathbb{E}[R]^{(t+1)} \geq \mathbb{E}[R]^{(t)}.
    \end{equation}
    This inequality holds because the optimization process aligns the planning agent with sub-goal sequences that facilitate better performance in the grounding agent.

\begin{lemma}\label{lem:ground}
    Optimizing the grounding agent while keeping the planning agent fixed leads to a non-decreasing expected reward.
\end{lemma}
    The grounding agent is optimized through supervised fine-tuning using pairs $(\boldsymbol{s}, \boldsymbol{a})$ generated by the planning agent. The corresponding optimization objective is:
    \begin{equation}
        \min\nolimits_{\pi_\mathrm{g}} \, \mathbb{E}_{(\boldsymbol{s}, \boldsymbol{a}) \sim \mathcal{S}} \left[ \mathcal{L}(\pi_{\mathrm{g}}(\boldsymbol{a} \mid \boldsymbol{s})) \right],
    \end{equation}
    where $\mathcal{L}$ denotes the loss function (i.e., cross-entropy loss). Minimizing this loss is equivalent to maximizing the log-likelihood of the correct tool invocation sequences:
    \begin{equation}
        \max\nolimits_{\pi_\mathrm{g}} \, \mathbb{E}_{(\boldsymbol{s}, \boldsymbol{a}) \sim \mathcal{S}} \left[ \log \pi_{\mathrm{g}}(\boldsymbol{a} | \boldsymbol{s}) \right].
    \end{equation}
    Since improved log-likelihood corresponds to reduced PPL and, consequently, higher reward~\cite{singh2023beyond}, it follows that:
    \begin{equation}
        \mathbb{E}[R]^{(t+1)} \geq \mathbb{E}[R]^{(t)}.
    \end{equation}
Hence, optimizing the grounding agent improves or maintains the expected reward when the planning agent is fixed.

From Lemma~\ref{lem:plan} and Lemma~\ref{lem:ground}, we establish that both optimization steps ensure non-decreasing expected rewards, i.e., $\mathbb{E}[R]^{(t+1)} \geq \mathbb{E}[R]^{(t)}$.
Additionally, the expected reward $\mathbb{E}[R]$ is upper-bounded due to the following reasons:
\begin{enumerate*}[label=(\roman*)]
    \item The reward function $R(\boldsymbol{s}, \boldsymbol{a})$ is bounded in practical scenarios; and
    \item The PPL has a lower bound.
\end{enumerate*}
Based on the \textit{Monotone Convergence Theorem}~\cite{Bibby_1974}, the non-decreasing and upper-bounded nature of $\{\mathbb{E}[R]^{(t)}\}^{\infty}_{t=1}$ ensures this sequence converges to a finite limit.
Thus, we derive the convergence of overall training process.

%% file: sections/04-experiment-set-up.tex
\section{Experimental Setup}\label{sec:exp_setup}

\input{table/task_eval}

\subsection{Benchmarks and Evaluation Metrics}
Following prior work~\cite{song2024agentbank,agentflan}, we evaluate \model under both held-in and held-out settings to evaluate its performance and generalization across diverse task types.
We consider a wide range of tasks, including mathematical reasoning, web interaction, and question answering.
As listed in Table~\ref{tab:test_tasks}, the held-in setting includes three tasks that are used during training: \textit{GSM8K}, \textit{StrategyQA}, and \textit{Mind2Web}; the held-out setting evaluates generalization on unseen tasks: \textit{SVAMP}, \textit{WebShop}, and \textit{HotpotQA}.
Evaluation metrics for each task are also reported in Table~\ref{tab:test_tasks}.
Following the recipe of baselins~\cite{lumos}, we define a set of action instructions (i.e., tool set $I$), covering common actions required for each task.
Details are provided in Appendix~\ref{app:tools}.

\input{table/main_results}
\subsection{Baselines}
We compare our \model with widely-used agent tuning methods, including:
(i) \textit{Agent Tuning}~\cite{zeng-etal-2024-agenttuning}, a multi-task tuning approach training LLMs on synthetic datasets comprising six tasks;
(ii) \textit{Agent-FLAN}~\cite{agentflan} employs a modular architecture that trains distinct single-agent capabilities through specialized parameter groups; and
(iii) \textit{Agent Lumos}~\cite{lumos}, a multi-agent training framework that separately fine-tunes models on datasets to obtain specialized agents.
Furthermore, we included GPT-3.5-Turbo and GPT-4 \cite{achiam2023gpt4} as strong single-agent baselines for comparison.

\subsection{Implementation Details}
To ensure a fair comparison with prior work, we adopt \texttt{Llama2-7b-hf} as the backbone LLM for both \model and baseline methods, following the official implementation of previous methods ~\cite{zeng-etal-2024-agenttuning,lumos}. To comprehensively evaluate our method across different LLMs, we additionally apply \model to two different model series with varying parameter scales: \texttt{Mistral-7B-Instruct-v0.2} and \texttt{Qwen2.5-14B}.
During the alignment process, we set the number of sampled subgoal sequences \(K\) to 15 and the number of training iterations to 2. The sampling temperature is set to 1.0 to encourage diversity in the generated subgoals.

We employ \texttt{DeepSeek-R1-Distill-Qwen-32B} as the critic model (denoted as \texttt{DS-Qwen-32B})  for verifying and correcting the generated tool-use action sequences. We further analyze the impact of using different critic models in Section~\ref{sec:critic_model}.
More details are provided in Appendix~\ref{app:exp}.

%% file: table/task_eval.tex
\begin{table}[t]
\centering
\resizebox{\linewidth}{!}{
\begin{tabular}{lcccc}
\toprule
\textbf{Task} & \textbf{Skill Dim.} & \textbf{\#Inst.} & \textbf{Metric} \\
\midrule
\multicolumn{4}{c}{\textbf{Held-in Tasks}} \\
\midrule
StrategyQA~\citep{yang2018hotpotqa} & QA & 300 & Exact Match \\

GSM8K~\citep{cobbe2021gsm8k} & Math & 1300 & Accuracy \\
Mind2Web~\citep{deng2023mind2web} & Web & 200 & Step Success Rate \\

\midrule
\multicolumn{4}{c}{\textbf{Held-out Tasks}} \\
\midrule
HotpotQA~\citep{geva2021strategyqa} & QA & 100 & Exact Match \\
SVAMP~\citep{patel2021svamp} & Math & 1000 & Accuracy \\
WebShop~\citep{yao2022webshop} & Web & 500 & Avg. Reward \\

\bottomrule
\end{tabular}
}
  \captionsetup{skip=5pt}
      \setlength{\belowcaptionskip}{-8pt}
\caption{The held-in and held-out tasks used to evaluate the agent capabilities of different LLMs.}
\label{tab:test_tasks}
\end{table}

%% file: table/main_results.tex
\begin{table*}[t]
\centering
\setlength\tabcolsep{5pt}
\begin{adjustbox}{width=1.95\columnwidth,center}
\begin{tabular}{l|c|ccc|c|ccc|c}
\toprule
\multirow{2}{*}{\textbf{Method}} & \multirow{2}{*}{\textbf{Base Model}} & \multicolumn{4}{c|}{\textbf{Held-in Tasks}} & \multicolumn{4}{c}{\textbf{Held-out Tasks}} \\
\cmidrule(l){3-6} \cmidrule(l){7-10} 
&  & GSK8K & Mind2Web & StrategyQA & Avg. & SVAMP & WebShop & HotpotQA  & Avg. \\
\midrule

\multicolumn{10}{c}{\textit{API-Based Agents}} \\
\midrule
GPT-4  & -  & 87.0  & 22.6 & 71.0 & 60.2 & 90.5 & 58.6 & 52.1 & 67.1 \\ 
GPT-3.5-Turbo & - & 65.0  & 21.7 & 58.0 & 48.2 & 81.0 & 62.4 & 24.0 & 55.8 \\ 
\midrule

\multicolumn{10}{c}{\textit{Llama Model Agents}} \\
\midrule
Llama-2-7B-Chat \textit{ }\textit{ }\textit{ } & Llama-2-7B  & 15.0  & 11.9 & 5.0 & 10.6 & 20.7 & 15.8 & 3.0 & 13.2  \\ 
Agent Tuning  & Llama-2-7B  & 14.0  & 10.6 & 49.0 & 24.5 & 35.3 & 59.8 & 10.0 & 35.0 \\ 
Agent Tuning  & Llama-2-13B  & 22.3 & 11.1 & \textbf{52.0} & 28.5 & 56.9 & \textbf{65.0} & 24.0 & 48.6 \\ 
Agent-FLAN & Llama-2-7B  &  28.5 & 16.9 & 48.0 & 31.1 & 39.2 & 55.9 & 12.0 & 35.7 \\
Agent Lumos  & Llama-2-7B  & 46.6 & 29.9 & 46.7 & 41.1 & 65.5 & 58.3 & 25.0 & 49.6  \\

\rowcolor{green!12}
\model  & Llama-2-7B  & \textbf{47.4} &\textbf{ 33.0} & \textbf{52.0} & \textbf{44.1} & \textbf{69.2} & 60.6 & \textbf{27.0} & \textbf{52.3} \\  
\midrule

\multicolumn{10}{c}{\textit{Mistral Model Agents}} \\
\midrule
Agent Lumos  & Mistral-7B-v0.2  & 46.4  & 33.8 & 49.3 & 43.2 & 61.9 & 58.7 & 27.0 & 49.2 \\ 

\rowcolor{green!12}
\model  & Mistral-7B-v0.2  & \textbf{48.2} & \textbf{34.7} & \textbf{56.0} & \textbf{46.3} & \textbf{73.7} & \textbf{59.0} & \textbf{28.0} & \textbf{53.6} \\

\midrule
\multicolumn{10}{c}{\textit{Qwen Model Agents}} \\
\midrule
Agent Lumos  & Qwen2.5-14B  & 81.7  & 31.8 & 49.3 & 54.3 & 85.5 & 64.7 & 27.0 & 59.1 \\ 

\rowcolor{green!12}
\model  & Qwen2.5-14B  & \textbf{82.4} & \textbf{32.6} & \textbf{55.3} & \textbf{56.8} & \textbf{87.4} & \textbf{65.8} & \textbf{28.0} & \textbf{60.4} \\

\bottomrule
\end{tabular}
\end{adjustbox}
\caption{Evaluation results of \model and baselines on both held-in and held-out tasks. The best results in each group are highlighted in \textbf{bold}.}\label{tab:main-results}
\end{table*}

%% file: sections/05-experiment.tex
\section{Experiment results}\label{sec:result}
\vspace{-1mm}

\subsection{Overall Performance}
\noindent\textbf{Held-in Tasks.} 
Table~\ref{tab:main-results} presents the evaluation results.
Compared with single-agent systems and independently trained multi-agent baselines, the \model achieves superior performance across three held-in tasks across different base models.
The \model with Llama-7B demonstrates an average improvement of 15.6\% compared to AgentTuning with \texttt{Llama-13B}.
{These improvements validate the effectiveness of our joint training framework, which tightly interconnects specialized agents to enhance overall task-solving performance.}

\noindent\textbf{Held-out Tasks.}
We further investigate the generalizability of our method in solving unseen tasks.
As illustrated in Table~\ref{tab:main-results}, our method achieves the highest performance compared to open-source baselines.
For example, the \model with \texttt{Mistral-7B} outperforms Lumos with an average performance improvement of 4.4\%.
An explanation for this improvement is that through iterative alignment in \model, the subgoals generated by the planning model align better with the preferences of the grounding models, and the grounding models also achieve a more accurate understanding of the generated subgoals.
This mutual understanding enhances the generalizability of the overall system when facing unseen tasks.

\noindent\textbf{Comparison with Closed-source Agents.}
Although our method is trained on 7B models like \texttt{Llama-7B}, it achieves about a 50\% performance improvement over GPT-4 on the Mind2Web task. 
This further validates the superiority of the \model in synergizing smaller open-source models to achieve competitive performance.

\input{table/ablation}

\begin{figure}[!t]
  \centering
  \includegraphics[width=0.47\textwidth]{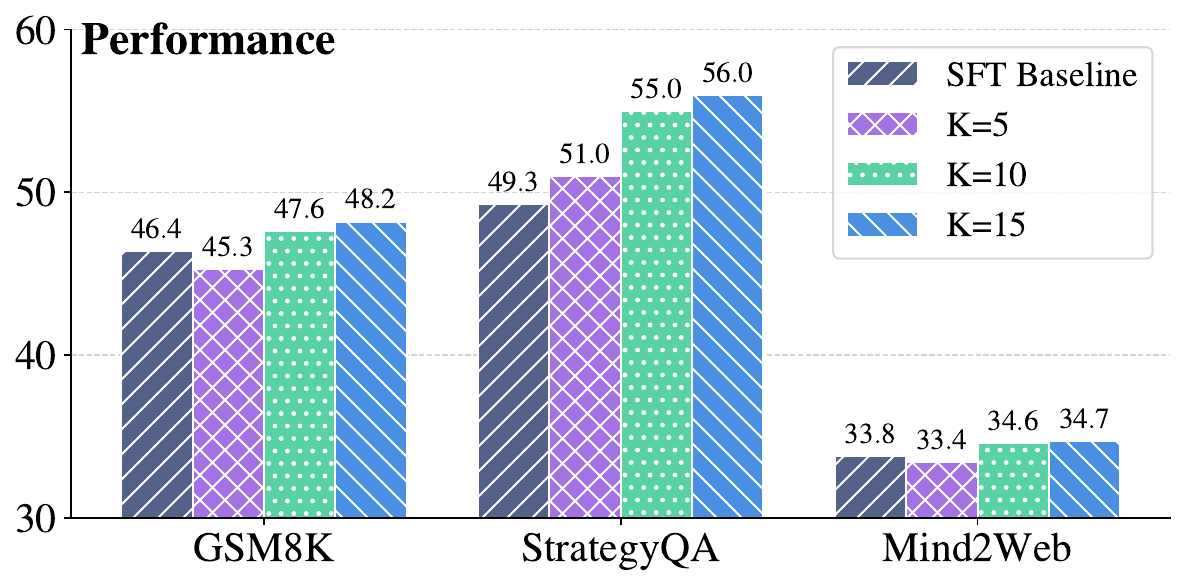} 
  \captionsetup{skip=4pt}
  \caption{Results of \model on three held-in tasks under different numbers of sampled subgoal sequences.}  
  \label{fig:bar_plot_1}  
\end{figure}

\subsection{Ablation Study}

To further analyze the contribution of each component in \model, we conduct an ablation study by removing individual components, including planning alignment (\textit{w/o stage 1}), grounding improvement (\textit{w/o stage 2}), and the critic model (\textit{w/o critic}), respectively.
As shown in Table~\ref{tab:ablation}, all variants exhibit substantial performance degradation, confirming the effectiveness of each component in our joint alignment framework. 

Besides, we highlight \textit{two key points}:
(1) the largest performance drop occurs in \textit{w/o stage 1}, highlighting the critical role of aligning the planning agent to generate coherent subgoals; and
(2) removing the critic model (\textit{w/o critic}) results in the second-largest performance drop, even lower than that caused by removing the grounding improvement (\textit{w/o stage 2}). This suggests that, without external feedback from the critic model, the system may suffer from significant negative updates, thereby validating the importance and rationality of incorporating a critic model.

\subsection{Analysis of the Capability Gap}

\input{table/capabilitygap}
To quantitatively validate our core hypothesis of a capability gap, we measured the perplexity of the grounding agent when generating correct actions. In this context, lower perplexity indicates better alignment, suggesting that the subgoals are more easily understood by the grounding agent. As shown in Table~\ref{tab:capability-gap}, the results provide direct evidence of this gap. The grounding agent consistently exhibits lower perplexity when processing subgoals from the planning agent in MOAT. In contrast, subgoals generated by the baseline without alignment result in higher perplexity, reflecting a distribution mismatch. The  reduction in perplexity demonstrates that our joint alignment process effectively harmonizes the agents, directly bridging this capability gap.

\vspace{-1mm}
\subsection{Hyperparameter Analysis}
\noindent\textbf{Analysis of Different Sample Numbers.} 
In our main experiments, we set the number of sampled responses $K$ to 15. To explore the impact of the sampling number $K$ on model performance, we further vary $K$ from 5 to 15 during the training of \texttt{Mistral-7B} at 2$th$ iteration.
As shown in Figure ~\ref{fig:bar_plot_1}, we observe a positive correlation between the sampling number and the overall performance.
We also identify a performance drop on the GSM8K and Mind2Web benchmarks when K=5. 
An explanation is that a smaller number of samples may fail to include high-quality subgoal sequences that align well with the grounding agent. In such cases, even the subgoal sequence with the highest reward may still be suboptimal or incorrect, thus negatively affecting training performance.

\begin{figure}[!t]
  \centering
  \captionsetup{skip=5pt}
  \includegraphics[width=0.47\textwidth]{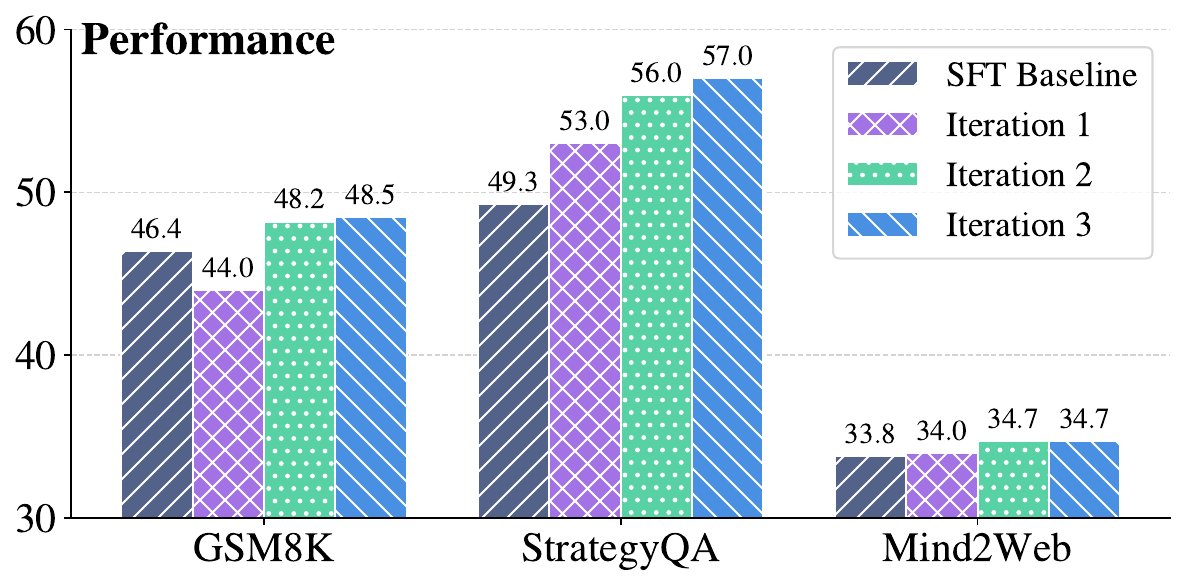} 
  \caption{Performance trends of \model (K=15) on held-in tasks as iterations increase. We use Accuracy / Exact Match / Step Success Rate as evaluation metrics for GSM8K / StrategyQA / Mind2Web datasets, respectively.}
  \label{fig:line_plot1}  
\end{figure}

\noindent\textbf{Analysis of Iteration Count.}
We further investigate how the iteration count impacts model performance using \texttt{Mistral-7B} with set K to 15.  As shown in Figure~\ref{fig:line_plot1}, the model’s performance improves gradually with the increasing number of iterations. However, by the third iteration, the performance gains become marginal. We suspect this is because, after several iterations, the planning and grounding agents gradually converge and reach a performance equilibrium, as discussed in Section~\ref{sec:therotical}.

\input{table/critic_ana}

\vspace{-1mm}
\subsection{Impact of Different Critic Model.}\label{sec:critic_model}

We use \texttt{DeepSeek-R1-Distill-Qwen-32B} as the default critic model in our framework to validate and refine the tool-use action sequences generated by the grounding agent.
To investigate the effect of the critic model’s capability, we conduct a comparative study using a stronger critic (GPT-4o) and a weaker one (\texttt{DeepSeek-R1-Distill-Qwen-14B}).
As shown in Table~\ref{tab:critic_result}, the results demonstrate an upward trend in task performance as the ability of the critic model increases.
However, we also observe that using a smaller, open-source model like \texttt{Qwen-14B} still yields competitive results, surpassing existing baselines by a notable margin.
We attribute this to the relatively simple nature of the critic’s task, i.e., verifying whether the predicted action sequence achieves the same effect as the ground-truth. Since both the prediction and reference are provided to the context of the critic model, this task requires less complex reasoning with simplified difficulty.
Therefore, while stronger critic models can further enhance performance, our framework remains robust and effective even when using smaller, fully open-source critics.

\begin{figure}[!t]
  \centering 
  \captionsetup{skip=5pt}
  \includegraphics[width=0.47\textwidth]{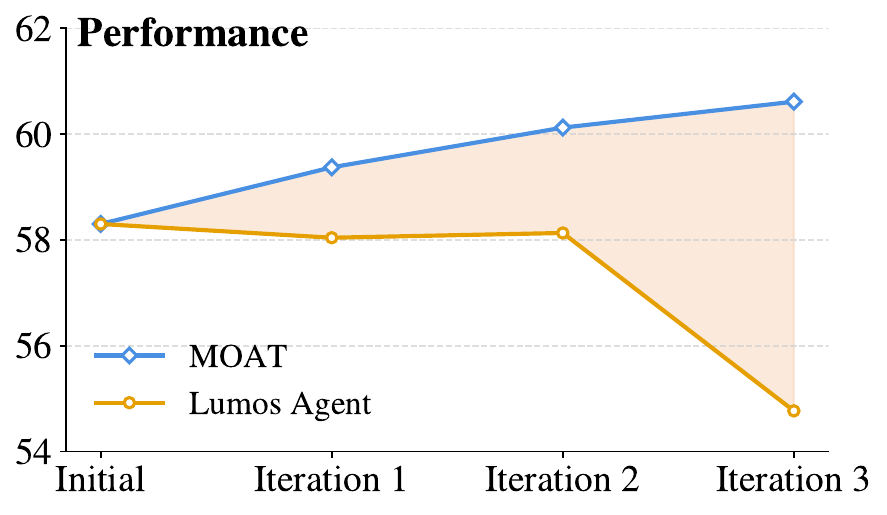} 
  \caption{Performance comparison on WebShop between MOAT and Lumos under equal training time. We use the Avg. Reward as evaluation metric.}
  \label{fig:line_plot2}  
\end{figure}

\subsection{Training Iteration Control Analysis}
A potential concern is that the observed performance gains from our iterative training strategy may stem merely from additional training epochs, rather than from the collaborative optimization of planning and grounding agents. To investigate this, we compare our approach with a baseline trained independently for the same total number of epochs.
Starting from a model trained for 2 epochs, we apply our iterative method for 1, 2, and 3 iterations, equivalent to 3, 4, and 5 total epochs.
We compare with the Lumos Agent baseline trained independently for the same number of epochs without inter-agent interaction.
As shown in Figure~\ref{fig:line_plot2}, the baseline struggles to consistently improve and even suffers from degradation due to overfitting. In contrast, our method shows consistent improvements, indicating that the gains stem from iterative co-training rather than extended training iterations.

\vspace{-1mm}
\subsection{Case Study}

We manually analyze the outputs of both the planning and grounding agents after training in \model. The results show that our \model effectively enhances the specialized expertise of both agents, as well as their adaptability. Concrete examples and detailed analysis are provided in Appendix~\ref{app:case}.

%% file: table/ablation.tex
\begin{table}[!t]
    \centering 
    \begin{adjustbox}{width=0.95\columnwidth,center}
    \setlength\tabcolsep{4pt}
    \begin{tabular}{@{}p{3cm}  c c c@{}} 
        \toprule
        \textbf{Method} & \textbf{Mind2Web} & \textbf{WebShop} & \textbf{Avg. $\Delta$} \\
        \midrule
        \rowcolor{gray!40} 
        
        \textit{Llama-2-7B} & & &\\
         Vanilla \model & 32.96 & 60.63  & 46.80 \\
        \textit{ -w/o stage 1} & 29.58$_{\downarrow 3.38}$ & 58.76$_{\downarrow 1.87}$  &  44.17$_{\downarrow 2.63}$ \\
    \textit{ -w/o stage 2} & 32.19$_{\downarrow 0.77}$ & 60.29$_{\downarrow 0.34}$  &  46.24 $_{\downarrow 0.56}$\\
    \textit{ -w/o critic} & 31.80$_{\downarrow 1.16}$ & 59.79$_{\downarrow 0.84}$  &  45.80 $_{\downarrow 1.00}$\\
        \bottomrule
    \end{tabular}
    \end{adjustbox}
  \captionsetup{skip=5pt}
    \setlength{\belowcaptionskip}{-6pt}
    \caption{Ablation study on two web datasets.}
    \label{tab:ablation}
\end{table}

%% file: table/capabilitygap.tex
\begin{table}[!t]
    \centering 
    \begin{adjustbox}{width=0.95\columnwidth,center}
    \setlength\tabcolsep{4pt}
    \begin{tabular}{@{}p{3cm}  c c c@{}} 
        \toprule
        \textbf{Task} & \textbf{Agent Lumos} & \textbf{MOAT} & \textbf{Reduction (↓\%)} \\
        \midrule
        \rowcolor{gray!40} 
        
        \textit{Mistral-7B} & & &\\
                  Math   & 3.53 & 2.56 & ${\downarrow27.48\%}$\\           
QA     & 1.41 & 1.40 &${\downarrow0.71\%}$\\

Web    & 5.71 & 5.02 &${\downarrow 12.08\%}$\\
        \bottomrule
    \end{tabular}
    \end{adjustbox}

    \caption{The perplexity comparison of the grounding agent across three tasks.}
    \label{tab:capability-gap}
\end{table}

%% file: table/critic_ana.tex

\begin{table}[!t]
    \centering 
    \begin{adjustbox}{width=0.98\columnwidth,center}
    \setlength\tabcolsep{4pt}
    \begin{tabular}{@{}l  cc c@{}} 
        \toprule
        \textbf{Model} & \textbf{Mind2Web} & \textbf{WebShop} & \textbf{Avg.} \\
        \midrule
        \rowcolor{gray!40} 
        
        \textit{Llama-2-7B} & & &\\
        \model \textit{w/ DS-Qwen-32B}  & 32.96 & 60.63 &  46.80 \\
        \model \textit{w/ GPT-4o}  & \textbf{34.45}$_{\uparrow4.52\%}$ & \textbf{60.78}$_{\uparrow0.25\%}$ & \textbf{47.62}$_{\uparrow1.75\%}$\\
        \model \textit{w/ DS-Qwen-14B}  & 31.80$_{\downarrow3.52\%}$ & 60.45$_{\downarrow0.30\%}$ &  46.13$_{\downarrow1.43\%}$ \\
        \bottomrule
    \end{tabular}
    \end{adjustbox}
    \caption{Model Performance on Mind2Web and  WebShop benchmarks using different critic models.} 
    \label{tab:critic_result}
\end{table}



%% file: sections/06-conclusion.tex
\vspace{-1mm}
\section{Conclusion}
\vspace{-1mm}

In this work, we present \model, a novel Joint Alignment Tuning framework designed to harmonize the collaboration between planning and grounding agents in LLM-based multi-agent systems. By iteratively optimizing the planning agent to generate subgoals that align with the grounding agent’s capabilities and enhancing the grounding agent’s adaptability to diverse subgoal sequences, \model effectively bridges the capability gap caused by independent training.
Both theoretical analysis and extensive experimental results demonstrate the superiority of \model.
We suggest that future work can explore the integration of visual models to expand the capability boundary of agents, as well as integrate more specialized agents, such as tool retrieval and reflection modules, to expand the versatility and efficiency of the system.

%% file: sections/appendix.tex
\section{Appendix}\label{sec:app}
\label{sec:appendix}

\subsection{Implementation Details} ~\label{app:exp}
We show more training details about our experiments. All our experiments are conducted on 6 × NVIDIA A800 (80GB) GPUs. 

For the initial fine-tuning stage, we use the public datasets provided by Lumos~\cite{lumos} and train two epochs with a learning rate of \(2 \times 10^{-5}\). And we set the maximum sequence length to 1024 and the batch size to 128. We also apply linear warmup for 3\% of the total training steps to adjust the learning rate.

For DPO training, we fine-tuned the model using the \texttt{accelerate} framework with DeepSpeed for optimized distributed training. We set batch size to 4 and  gradient accumulation to 8. The learning rate is set to \(4 \times 10^{-7}\) with a cosine learning rate scheduler. And We set the maximum sequence length to 1024. Additionally, we leveraged the \texttt{TRL} library \footnote{\url{https://github.com/huggingface/trl}} to facilitate the training of reinforcement learning-based models.
Meanwhile, we filter out data samples where the reward difference between \( s_w \) and \( s_l \) is less than 0.1 for stability during DPO training.

For grounding agent improving training, we implement training over two epochs with a learning rate of \(2 \times 10^{-5}\) and a batch size 128 the same with initial tuning. 
At the same time, we mix these with the initial data in a 1:1 ratio to prevent the model from forgetting prior knowledge.

\input{table/iter_op}

\subsection{Case Study} ~\label{app:case}
As illustrated in Table~\ref{app:case}, the case studies evaluate the responses generated by our \model and the independent training method. Our findings show that through joint alignment tuning, the models are able to align their capabilities. Specifically, for the given case, we observe that the independently trained method struggles with subgoal decomposition in planning agent, making it difficult for grounding agent to resolve, leading to a failure in solving the task. However, after the joint alignment training, the capability gap is addressed, allowing the planning agent to generate subgoals that are easier for the grounding agent to understand. Consequently, the grounding agent successfully produced the correct action sequence.

\input{table/case}

\subsection{Prompts}\label{app:prompt}
We provide our prompts used in the multi-agent system. The prompt of the planning agent to generate the subgoal sequences is in Table~\ref{tab:prompt-planning}. The prompt of the grounding agent to generate the action sequences is in Table~\ref{tab:prompt-grounding}. The prompt of the critic model to correct the action sequences generated by grounding agents is in Table~\ref{tab:prompt-critic}. 
\input{table/prompts}

\subsection{Action Interfaces and Execution Tools for Complex Interactive Tasks}
\label{app:tools}

\input{table/tools}

For each defined action in the action interfaces, a corresponding backend execution tool is provided to enable the implementation of that action.
Our setup follows the approach described in \citet{yin2024agentlumosunifiedmodular}. We have adopted the same configuration to ensure comparability between our work and theirs.

As shown in Table~\ref{table:appendix-tools-qa}, for QA tasks, we use Wikipedia and Google Search APIs to find relevant knowledge about entities. Additionally, we use a semantic matching model, \texttt{dpr-reader-multiset-base}\footnote{\url{https://huggingface.co/facebook/dpr-reader-multiset-base}.}, employed in Dense Passage Retrieval (DPR)~\citep{karpukhin-etal-2020-dense}, to retrieve paragraphs based on the query. Following the approach from ReWOO~\citep{xu2023rewoo}, we also utilize GPT-series models as a straightforward QA tool to respond to queries based on the retrieved knowledge or prior interactions.

In Table~\ref{table:appendix-tools-web}, web tasks involve real mouse and keyboard operations such as typing, clicking, and selecting HTML tags. To identify the appropriate HTML tags to operate on, we use a DeBERTa model\footnote{\url{https://huggingface.co/osunlp/MindAct_CandidateGeneration_deberta-v3-base}.} that ranks and retrieves relevant tags based on the current action, as seen in the AgentBench evaluation.

As illustrated in Table~\ref{table:appendix-tools-math}, WolframAlpha API~\footnote{\url{https://www.wolframalpha.com/}.} serves as the main tool for mathematical tasks, as it is capable of executing a wide range of mathematical functions, including formula computation and equation solving. For more advanced math operations like sorting, we leverage OpenAI Codex~\citep{chen2021evaluating} to generate short code snippets for execution.

For the unseen task WebShop, the actions include \texttt{Search}, \texttt{FeatureRetrieve}, \texttt{Pick}, and \texttt{Click}. The \texttt{Search} and \texttt{Click} actions are implemented using the embedded features provided in the official WebShop virtual environment\footnote{\url{https://github.com/princeton-nlp/WebShop}.} following \cite{liuagentbench}. Meanwhile, \texttt{FeatureRetrieve} and \texttt{Pick} rely on the \texttt{dpr-reader-multiset-base}, which helps select the most relevant items and their features based on the query.

%% file: table/iter_op.tex
\begin{algorithm}[!t]
    \SetAlgoLined
    \KwIn{The number of iterations $N_r$, the number of samples $K$, the tasks set $\mathcal{D}$,  the set of available tools $I$, 
    the critic model \textsc{Critic} ,
    planning agent $\pi_{p}$ ,
    grounding agent $\pi_{g}$.}
    \BlankLine
    \For{iteration t = 1... $N$}{
        \emph{\# \textcolor{blue}{Initialize the training dataset }} \\
        $\mathcal{D}_{p} \leftarrow \phi, \mathcal{D}_{g} \leftarrow \phi$
        $\mathcal{R} \leftarrow \phi$
        \BlankLine
        \emph{\# \textcolor{blue}{Planning Agent Optimization}} \\
        \For{each task $x_i \in \mathcal{D}$}{
        \emph{\# \textcolor{blue}{sample $K$ response}} \\
            \For{$j\leftarrow 1$ \KwTo $K$}{
                $\boldsymbol{s}_{i,j} \sim \pi_{p}^{t-1}(x_i)$ \\
                \emph{\# \textcolor{blue}{$\tilde{\boldsymbol{a}}_{i,j}$ denotes correct actions}} \\
                $\boldsymbol{r}_{i,j} \leftarrow \mathrm{PPL}_{\pi_{g}^{t-1}}       (\tilde{\boldsymbol{a}}_{i,j} \vert x_i, I,\boldsymbol{s}_{i,j})$ \\
                $\mathcal{D}_{p,i} \leftarrow \mathcal{D}_{p,i} \cup\boldsymbol{s}_{i,j}$\\
                $\mathcal{R}_{p,i} \leftarrow \mathcal{R}_{p,i} \cup\boldsymbol{r}_{i,j}$
            }
        }
        \BlankLine
        $\mathcal{S}_{win} \leftarrow \mathcal{D}_{p}[\operatorname{argmin}(\mathcal{R}, axis=1)] $ \\
        $\mathcal{S}_{lose} \leftarrow \mathcal{D}_{p}[\operatorname{argmax}(\mathcal{R}, axis=1)] $ \\
        \BlankLine
        $\pi_{p}^{t} \leftarrow  \pi_{p}^{t-1} - \nabla_{\pi_{p}}\mathcal{L}_{\mathrm{DPO}}(\mathcal{D}, \mathcal{S}_{win}, \mathcal{S}_{loss})$
        \BlankLine
        \emph{\# \textcolor{blue}{Grounding Agent Optimization}} \\
        \For{each task $x_i \in D$}{
            \For{each $\boldsymbol{s}_{i,j} \in \mathcal{D}_{p} $}{
                $\boldsymbol{a}_{i,j} \sim \pi_{g}^{r-1}(\boldsymbol{a}_{i,j}\vert x_i, I, \boldsymbol{s}_{i,j}) $ \\ 
                $ \hat{\boldsymbol{a}}_{i,j} \leftarrow \textrm{Critic}(\boldsymbol{a}_{i,j}, \tilde{\boldsymbol{a}}_{i,j}) $ \\
                $\mathcal{D}_{g} \leftarrow \mathcal{D}_{g}\cup (x,I,\boldsymbol{s}_{i,j},\hat{\boldsymbol{a}}_{i,j})$
            }
        }
        $\pi_{g}^{t} \leftarrow \pi_{g}^{t-1}  - \nabla_{\pi_{g}}\mathcal{L}_{\mathrm{SFT}}(\mathcal{D}_g)$
    } 
    \KwOut{$\pi_{p}^{N}$, $\pi_{g}^{N}$}
\caption{The proposed multi-agent joint optimization, which iteratively aligns the planning and grounding agents, improving the holistic task-solving performance.}
\label{alg:iterative_optimization}
\end{algorithm}

%% file: table/case.tex










\begin{table*}[htbp]
\small \centering
\setlength\tabcolsep{20pt}
\begin{tabular}{@{}p{16cm}@{}}
\toprule
\textbf{Case Study} \\
\hline

\\
Question:\\
Richard lives in an apartment building with 15 floors.  Each floor contains 8 units, and 3/4 of the building is occupied.  What's the total number of unoccupied units In the building? \\ \\
\hline
\\
\textbf{\model:} \\ \\
\textbf{Subgoals:}\\
Subgoal 1: Calculate the total number of units in the entire building.;   \\
Subgoal 2: Calculate the number of floors that are occupied.;  \\
Subgoal 3: Calculate the total number of units that are occupied.;  \\
Subgoal 4: Calculate the total number of unoccupied units in the building.\\ \\

\textbf{Actions:}\\
R1 = Calculator(15 * 8);  \\
R2 = Calculator(15 * 3/4);  \\
R3 = Calculator(R2 * 8);  \\
R4 = Calculator(R1 - R3) \\ \\

\textbf{Answer:} \\
30 \\
\\
\hline \\
\textbf{Independent Training Method:} \\ \\
\textbf{Subgoals:}\\
Subgoal 1: Calculate the total number of units in the building.;  \\
Subgoal 2: Calculate the number of units that are unoccupied.; \\
Subgoal 3: Calculate the total number of unoccupied units in the building. \\ \\

\textbf{Actions:}\\
R1 = Calculator(15 * 8);  \\
R2 = Calculator(R1 * 3/4);  \\
R3 = Calculator(R2 * 2) \\ \\

\textbf{Answer:} \\
180
\\
\bottomrule
\end{tabular}
\caption{A case study in the GSM8K test dataset.}
\label{tab:prompt-low}
\end{table*}

%% file: table/prompts.tex
\begin{table*}[htbp]
\small \centering
\setlength\tabcolsep{4pt}
\begin{tabular}{@{}p{16cm}@{}}
\toprule
\textbf{Prompt to generate subgoal sequences}
\\
\hline
\\
Please provide a reasonable subgoal-based plan to solve the given task. \\
\\
Task:\{TASK\} \\
\\
\bottomrule
\end{tabular}
\caption{The prompt of planning agent to generate subgoal sequences.}
\label{tab:prompt-planning}
\end{table*}

\begin{table*}[htbp]
\small \centering
\setlength\tabcolsep{4pt}
\begin{tabular}{@{}p{16cm}@{}}
\toprule
\textbf{Prompt to generate action sequences}
\\
\hline
\\
Please ground the given subgoal to corresponding executable actions for solving the given task. The grounded actions must be the one in available action list. \\
\\
The available action list is:\{ACTION\_LIST\} \\
\\
Task:\{TASK\} \\
\\
Subgoals to be grounded:\{SUBGOALS\}
\\
\bottomrule
\end{tabular}
\caption{The prompt of grounding agent to generate action sequences.}
\label{tab:prompt-grounding}
\end{table*}

\begin{table*}[htbp]
\small \centering
\setlength\tabcolsep{4pt}
\begin{tabular}{@{}p{16cm}@{}}
\toprule
\textbf{Prompt to correct action sequences}
\\
\hline
\\
Given a task and a corresponding series of subgoals and their corresponding actions that may be incomplete, your task is to judge whether the subgoals and actions can reached a final answer or conclusion for the problem. \\
The grounded actions must be the one in available action list.The available action list is \{ACTION\_LIST\} 
\\
If the actions can reached a final answer, you  should directly output "Final answer reached". Otherwise, you should give corrections to the original subgoals and their corresponding actions. It is not necessary to be similar to the original subgoals and actions.\\
\\
Task:\{TASK\} \\
Reference subgoals: \{REF\_SUBGOALS\} \\
Reference actions: \{REF\_ACTIONS\} \\
Judged subgoals: \{SUBGOALS\} \\
Judged actions: \{ACTIONS\} \\
\\
Your output should follow the format: \\
If can reached a final answer, directly output "Final answer reached".
Else, output corrected subgoals and actions following this format:  \\
Corrected Subgoals: <series of subgoals to complete the task  in one line, Each Subgoal begins with Subgoal idx> \\
Corrected Actions: <corresponding actions in one line> \\
\\
\bottomrule
\end{tabular}
\caption{The prompt of critic model to correct action sequences.}
\label{tab:prompt-critic}
\end{table*}

%% file: table/tools.tex
\begin{table*}[t]
\centering
\vspace{-0em}
\begin{subtable}[t]{\textwidth}
\centering
\scalebox{0.7}{
\begin{tabular}{cccc}
\toprule
\textbf{Task Type}                  & \textbf{Action Types}   & \textbf{Function Descriptions} & \textbf{Tools} \\ \midrule
\multirow{7}{*}{QA} & \texttt{KnowledgeQuery(Entity) -> Knowledge}    & Query the entity knowledge   & Wikipedia, Google Search              \\ \cmidrule{2-4}
& \makecell{\texttt{ParagraphRetrieval(Knowledge, Query)} \\ \texttt{-> Paragraphs}}    & \makecell{Retrieve relevant paragraphs \\ based on the query}   & \texttt{dpr-reader-multiset-base}              \\ \cmidrule{2-4}
& \texttt{QA(Context, Query) -> Answer}    & \makecell{Answer the query based on \\ the provided context}   & GPT-series/open LLMs              \\ \cmidrule{2-4}
& \texttt{Calculator(Expression) -> Value}  & Calculate given mathematical expressions   & WolframAlpha \\ \bottomrule
\end{tabular}
}
\caption{Actions used in complex QA tasks.}
\label{table:appendix-tools-qa}
\end{subtable}
\vspace{9pt}

\begin{subtable}[t]{\textwidth}
\centering
\scalebox{0.69}{
\begin{tabular}{cccc}
\toprule
\textbf{Task Type}                  & \textbf{Action Types}   & \textbf{Function Descriptions} & \textbf{Implementation} \\ \midrule
\multirow{6}{*}{Web} & \texttt{Click(Env, Query) -> Tag}    & \makecell{Locate the tag to be clicked based on the query}   & \multirow{6}{*}{\makecell{HTML Simulator}}              \\ \cmidrule{2-3}
& \texttt{Type(Env, Query, Text) -> Tag, Text}    & \makecell{Locate the relevant tag based on the query \\ and output the typed text}    &               \\ \cmidrule{2-3}
& \texttt{Select(Env, Query, Text) -> Tag, Text}    & \makecell{Locate the relevant tag based on the query \\ and output the selected option}   &               \\
\bottomrule
\end{tabular}
}
\caption{Actions used in web tasks.}
\label{table:appendix-tools-web}
\end{subtable}
\vspace{9pt}

\begin{subtable}[t]{\textwidth}
\centering
\scalebox{0.7}{
\begin{tabular}{cccc}
\toprule
\textbf{Task Type}                  & \textbf{Action Types}   & \textbf{Function Descriptions} & \textbf{Implementation} \\ \midrule
\multirow{8}{*}{Math}      & \texttt{Calculator(Expression) -> Value} & Calculate mathematical expressions   & \multirow{6}{*}{WolframAlpha}     \\ \cmidrule{2-3}
& \texttt{SetEquation(Expression) -> Equation} & Set equations based on the given expression &                    \\ \cmidrule{2-3}
& \texttt{SolveEquation(Equation) -> Solutions}      & Solve the system of equations  &                  \\ \cmidrule{2-3}
& \texttt{Define(Variable) -> Variable} & Define a variable &                    \\ \cmidrule{2-3}
& \texttt{SolveInequality(Inequality) -> Solutions}      & Solve the inequality  &                  \\ \cmidrule{2-4}
& \texttt{Code(Function\_Description) -> Code} & Generate code for mathematical functions & \texttt{gpt-3.5-turbo}        \\ \cmidrule{2-4}
& \texttt{Count(List) -> Number}      & Count the number of elements in a list  & Python                 \\ \bottomrule
\end{tabular}
}
\caption{Actions used in math tasks.}
\label{table:appendix-tools-math}
\end{subtable}

\caption{Action interfaces and execution module implementations for complex interactive tasks.}
\label{table:appendix-tools}
\end{table*}